\documentclass[lettersize,journal]{IEEEtran}

\usepackage{amsmath,amssymb,amsfonts}
\usepackage{algorithmic}
\usepackage{algorithm}
\usepackage{array}
\usepackage[caption=false,font=normalsize,labelfont=sf,textfont=sf]{subfig}
\usepackage{textcomp}
\usepackage{stfloats}
\usepackage{url}
\usepackage{verbatim}
\usepackage{graphicx}
\usepackage{cite}
\usepackage{xcolor}
\usepackage{booktabs}
\usepackage{microtype}
\usepackage{bm}
\usepackage{multirow}
\usepackage[hidelinks]{hyperref}

\graphicspath{{figures/}{./}}
\providecommand{\R}{\mathbb{R}}
\providecommand{\C}{\mathbb{C}}
\providecommand{\mD}{\text{mD}}

\begin{document}

\IEEEoverridecommandlockouts

\title{StarBOA: Real-Time Mamba State-Space Unrolling for Sparse Radar Micro-Doppler in ISAC Networks}

\author{Mustafa Bora \protect\c{C}elik, Ceren \protect\c{C}elik, and Orhan Gazi%
\thanks{This work has been submitted to the IEEE for possible publication. Copyright may be transferred without notice, after which this version may no longer be accessible.}%
\thanks{Corresponding author: Mustafa Bora \c{C}elik (e-mail: \protect\href{mailto:mustafa.celik1@std.ankaramedipol.edu.tr}{mustafa.celik1@std.ankaramedipol.edu.tr}).}%
\thanks{Code and reproducible models are publicly available at \texttt{https://github.com/BOA-clk/StarBOA}.}}

\markboth{Preprint. Under review.}%
{\protect\c{C}elik \textit{et al.}: StarBOA: Real-Time Mamba State-Space Unrolling for Sparse Radar Micro-Doppler in ISAC Networks}

\maketitle

\begin{abstract}
In Integrated Sensing and Communications (ISAC), radar sensing must operate under chirp subsampling with up to 90\% missing data. An attention-based baseline, limited to a 52~ms buffer, collapses toward maximum uniform entropy ($H=2.584$ bits) as sparsity increases, failing to capture long-range gait-cycle context. We propose StarBOA, which replaces attention with a causal Mamba state-space model that updates incrementally on a per-window basis without re-scanning past reconstructions. By maintaining a persistent state, StarBOA integrates over $100\times$ more temporal history at no additional per-step computational cost. StarBOA outperforms the baseline's published results across all sparsity levels, with SSIM gains increasing from $+0.0379$ at 50\% missing data to $+0.2472$ at 90\%. Each window is processed in 1.53~ms with zero lookahead, demonstrating efficient causal reconstruction under extreme chirp subsampling.
\end{abstract}

\begin{IEEEkeywords}
ISAC, Micro-Doppler Radar, Compressive Sensing, Algorithm Unrolling, State-Space Models, Mamba, Real-Time Edge Processing.
\end{IEEEkeywords}

\section{Introduction}
\label{sec:intro}
\IEEEPARstart{I}{SAC} is a foundational technology for next-generation $6\text{G}$ cellular systems, merging high-rate wireless data transmission with radar sensing over shared spectrum and hardware platforms \cite{wymeersch2021integration, liu2020joint, sparcs_dataset}. To maximize communication data throughput, base stations surrender up to $90\%$ of transmission time-frequency resource blocks to downlink data packets. As a consequence, radar sensing is confined to scarce, intermittently gathered chirp pulses ($10\%$ duty cycle). As highlighted in recent compressive sensing radar studies \cite{mazzieri2024attention, chen_micro_doppler}, reconstructing high-fidelity micro-Doppler ($\mD$) spectrograms from such heavily undersampled measurements is vital: pulse dropouts create severe velocity aliasing and noise artifacts that obscure subtle kinematic Doppler signatures essential for human activity recognition (HAR).

To reconstruct sparse radar returns, deep algorithm unrolling has emerged as a powerful paradigm, unrolling iterative compressive sensing solvers into neural network layers \cite{gregor2010learning, monga2021algorithm}. In this context, the Single Thresholding with Attention
Refinement (STAR) framework \cite{mazzieri2024attention} combines Learned Iterative Hard Thresholding (LIHT) with temporal attention and solution refinement, showing promising reconstruction under mild subsampling ($50\%$ missing chirps).

\subsection{Limitations of Attention Under Severe Sparsity}
Despite its effectiveness under mild subsampling, STAR's non-parametric attention mechanism exhibits fundamental limitations as pulse missingness escalates. STAR computes correlation weights directly in the frequency domain without learnable projection matrices: $\alpha_i[t] = \text{Softmax}(\frac{1}{\sqrt{K}} \mathbf{y}[t-i]^T \tilde{\mathbf{y}}[t])$ for $i \in \{1, \dots, N_p\}$. Under severe $90\%$ missingness, the candidate spectrum $\tilde{\mathbf{y}}[t]$ is dominated by compressive sensing noise ($\text{SNR} < 0\text{ dB}$). Consequently, inner products with past recovered spectra lose selective discrimination, flattening attention weights across the historical buffer into an unweighted uniform moving average ($\alpha_i[t] \approx 1/N_p = 1/6$). Rather than retrieving coherent gait patterns, attention simply averages multiple corrupted spectrogram windows, diffusing Doppler energy and blurring subtle limb trajectories into indistinct veils. Moreover, its $N_p=6$ buffer spans only $\approx 52\text{ ms}$ ($6 \times 8.64\text{ ms}$)—far short of a full gait cycle ($\approx 1.0\text{ s}$).

\subsection{Contributions: Continuous-Time State-Space Dynamics}
To address this challenge, we introduce \textbf{StarBOA} (Fig.~\ref{fig:arch}):
\begin{itemize}
    \item \textbf{Mamba-Based Recurrent Context Modeling:} We replace STAR's non-parametric attention mechanism with a selective continuous-time state space \cite{gu2023mamba}. Rather than re-scanning a fixed buffer, StarBOA carries a recurrent state with constant ($O(1)$) per-step update cost regardless of history length, integrating context across $128$ windows in our experiments—extendable further—versus STAR's fixed $6$-window buffer.
    \item \textbf{Root-Cause Entropy Collapse Analysis:} We reveal that under $90\%$ chirp missingness, STAR's dot-product attention collapses to maximum uniform entropy ($H = 2.5840\text{ bits} \approx \log_2(6)$), explaining why attention degrades to moving-average blur while recurrent state-space models remain structurally resilient against instantaneous noise.
    \item \textbf{Decoupled Front-End Benchmark Superiority:} On the public DISC $60\text{ GHz}$ benchmark, in decoupled evaluation with a frozen LIHT front-end, StarBOA outperforms STAR-Attention across all sparsity tiers ($50\%$, $75\%$, $90\%$), lifting kinematic structure correlation above $s > 0.888$ even at extreme sparsity.
    \item \textbf{Full Pipeline Verification Across All Sparsity Tiers:} Against STAR's officially reported results, StarBOA surpasses it at every tested sparsity level, with the SSIM gain widening from $+0.0379$ at $50\%$ to $+0.2472$ at $90\%$ sparsity.
    \item \textbf{Incremental Streaming Without History Reprocessing:} Unlike attention, which re-compares each new window against a stored frame buffer, StarBOA's recurrent state updates in constant time with no history reprocessing. Combined with a $10\%$ sensing duty cycle, this sustains real-time ISAC streaming ($1.53\text{ ms}$/window, zero lookahead) while freeing $90\%$ of time-frequency resources for communications.
\end{itemize}

\section{Preliminaries: The STAR Framework}
\label{sec:preliminaries}
Because StarBOA directly builds upon the unrolled architecture introduced in \cite{mazzieri2024attention}, we briefly summarize STAR's three cascaded blocks.

\subsection{Channel Model and Incomplete Sampling}
Under ISAC slot sharing, Channel Impulse Response (CIR) samples are gathered in slow-time windows of $K$ chirps with step shift $\delta$. Due to packet transmission patterns, only a subset $M_t \ll K$ of samples is gathered at window $t$. Let $\mathbf{h}[t] \in \C^{M_t}$ denote the incomplete measurement vector:
\begin{equation}
\mathbf{h}[t] = \mathbf{M}_t \mathbf{F}_K \mathbf{z}[t] + \mathbf{n}[t],
\label{eq:cs_model}
\end{equation}
where $\mathbf{M}_t \in \R^{M_t \times K}$ selects the sampled indices, $\mathbf{F}_K$ is the inverse Fourier dictionary, $\mathbf{z}[t] \in \C^K$ is the unknown sparse Doppler spectrum, and $\mathbf{n}[t]$ is noise \cite{mazzieri2024attention, sparcs_dataset}.

\subsection{The Three-Block STAR Pipeline}
The Single Thresholding with Attention Refinement (STAR) framework \cite{mazzieri2024attention} recovers sequential Doppler spectra through three cascaded processing blocks:
\begin{enumerate}
    \item \textbf{Block A (Single-Layer LIHT Module):} To avoid multi-iteration solver delay \cite{eldar2012compressed}, STAR unrolls a single Iterative Hard Thresholding step into a learnable physical layer:
    \begin{align}
    \mathbf{z}^{(0)}[t] &= \mathcal{H}_\Omega \left( \frac{1}{\mu} \mathbf{W}^T \mathbf{h}[t] \right), \\
    \mathbf{z}[t] &= \mathcal{H}_\Omega \left( \left(\mathbf{I} - \frac{1}{\mu} \mathbf{W}^T \mathbf{W}\right) \mathbf{z}^{(0)}[t] + \frac{1}{\mu} \mathbf{W}^T \mathbf{h}[t] \right),
    \label{eq:liht_star}
    \end{align}
    where $\mathcal{H}_\Omega$ preserves the $\Omega$ largest components, $\mu = 20$ is the inverse step size, and $\mathbf{W} \in \R^{2M_t \times 2K}$ is a dictionary initialized as $\mathbf{W} = R(\mathbf{M}_t \mathbf{F}_K)$ via real-complex mapping $R(\cdot)$ \cite{mazzieri2024attention}. The coarse spectrum is $\tilde{\mathbf{y}}[t] = [\mathbf{I}_K \; \mathbf{I}_K] \mathbf{z}[t]^2 \in \R^K$.
    \item \textbf{Block B (Attention Mechanism):} To capture temporal correlation, STAR compares candidate $\tilde{\mathbf{y}}[t]$ against $N_p=6$ past recovered spectra $\mathbf{Y}[t] = [\mathbf{y}[t-1], \dots, \mathbf{y}[t-N_p]]^T \in \R^{N_p \times K}$ using non-parametric scaled dot-product attention (without learnable projection matrices) to extract a context feature vector $\mathbf{a}[t]$:
    \begin{equation}
    \mathbf{a}[t] = \mathbf{Y}[t]^T \text{Softmax}\left( \frac{1}{\sqrt{K}} \mathbf{Y}[t] \tilde{\mathbf{y}}[t] \right) \in \R^K.
    \label{eq:attn_star}
    \end{equation}
    \item \textbf{Block C (Solution Refinement Module):} The coarse spectrum $\tilde{\mathbf{y}}[t]$ is refined based on the context feature vector $\mathbf{a}[t]$ via additive transformation and multiplicative gating:
    \begin{equation}
    \mathbf{y}[t] = \left( \tilde{\mathbf{y}}[t] + \text{ReLU}(\mathbf{U}\mathbf{a}[t] + \mathbf{b}) \right) \odot \sigma(\mathbf{V}\mathbf{a}[t]),
    \label{eq:refine_star}
    \end{equation}
    where $\mathbf{U}, \mathbf{V} \in \R^{K \times K}$ and $\mathbf{b} \in \R^K$ are learnable weights.
\end{enumerate}
\begin{figure}[t]
\centering
\includegraphics[width=0.85\columnwidth]{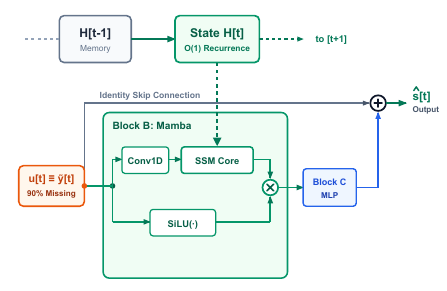}
\caption{Causal StarBOA architecture and recurrent state memory propagation across slow-time radar windows. Rather than matching against a noisy historical buffer, the compact $O(1)$ recurrent state $\mathbf{H}_t$ continuously accumulates sequential context across severe $90\%$ missing chirp bursts, enabling real-time streaming reconstruction with strictly $0.00\text{ ms}$ lookahead.}
\label{fig:arch}
\end{figure}
\textbf{Preserved Components:} We retain Blocks A and C unchanged, replacing only the vulnerable non-parametric attention (Block~B) with continuous-time selective state-space modeling.

\section{Proposed StarBOA Architecture}
\label{sec:system}

\subsection{Sequential State-Space Context Modeling (Block B)}
As illustrated in Fig.~\ref{fig:arch}, StarBOA replaces STAR's dot-product attention mechanism (Block~B) with a continuous-time selective state-space model \cite{gu2023mamba}. Instead of computing non-parametric affinity across a limited buffer of noisy frames, Mamba assimilates continuous historical information to predict and reconstruct the missing components in newly arriving slow-time windows.

For each feature channel $d \in \{1, \dots, D\}$ ($D=64$), the hidden state $\mathbf{h}_d(t) \in \R^N$ ($N=16$) evolves continuously according to:
\begin{equation}
\begin{aligned}
\dot{\mathbf{h}}_d(t) &= \mathbf{A}_d \mathbf{h}_d(t) 
+ \mathbf{B}_{d,t} x_{\text{conv},d}(t), \\
y_d(t) &= \mathbf{C}_{d,t} \mathbf{h}_d(t) 
+ D_d x_{\text{conv},d}(t),
\end{aligned}
\label{eq:continuous_ssm}
\end{equation}

where $\mathbf{A} \in \R^{D \times N}$ is the diagonal state transition matrix with $\mathbf{A}_{d, n} < 0$ ensuring numerical stability, and $\mathbf{B}_{d, t}, \mathbf{C}_{d, t} \in \R^N$ are input-dependent selective control vectors.

\textit{Discretization and Selective Recurrence:} At each slow-time window $t$, the coarse normalized spectrum $\mathbf{u}_t \equiv \tilde{\mathbf{y}}[t] \in \R^D$ ($D=K=64$ Doppler bins) produced by Block~A is projected into two parallel branches: an SSM branch $\mathbf{x}_t$ and a gating branch $\mathbf{z}_t$. Input $\mathbf{x}_t$ is processed by a depthwise causal 1D convolution ($d_{\text{conv}}=4$) with SiLU activation: $\mathbf{x}_{\text{conv}, t} = \text{SiLU}(\text{Conv1D}(\mathbf{x}_t))$, buffering the prior 3 steps in streaming mode. The input-dependent timescale $\Delta_t \in \R^D$ and selective vectors $\mathbf{B}_t, \mathbf{C}_t \in \R^N$ are dynamically generated from $\mathbf{x}_{\text{conv}, t}$, with $\Delta_t = \text{clip}(\text{Softplus}(\mathbf{W}_\Delta \mathbf{x}_{\text{conv}, t} + \mathbf{b}_\Delta), 10^{-4}, 0.1)$. Discretizing via first-order Euler approximation yields:
\begin{equation}
\bar{\mathbf{A}}_t = \exp\left(\Delta_t \mathbf{A}\right), \quad \bar{\mathbf{B}}_t = \Delta_t \mathbf{B}_t.
\label{eq:zoh}
\end{equation}
In online edge streaming, the recurrent state update is strictly causal ($0.00\text{ ms}$ lookahead):
\begin{align}
\mathbf{H}_t &= \bar{\mathbf{A}}_t \odot \mathbf{H}_{t-1} + \mathbf{x}_{\text{conv}, t} \otimes \bar{\mathbf{B}}_t, \label{eq:causal_mamba} \\
\mathbf{m}_t &= \mathbf{W}_{\text{out}} \left( \left( \sum_{n=1}^N \mathbf{H}_{t, \cdot, n} C_{t, n} + \mathbf{D} \odot \mathbf{x}_{\text{conv}, t} \right) \odot \text{SiLU}(\mathbf{z}_t) \right).
\end{align}

\subsection{Residual Solution Refinement (Block C)}
In STAR \cite{mazzieri2024attention}, Block~C required a separate multiplicative Sigmoid gate because attention provided no internal gating. In contrast, since Mamba already incorporates continuous SiLU gating within Block~B, StarBOA implements Block~C directly on top of the gated context $\mathbf{m}_t$ via a zero-initialized residual Multi-Layer Perceptron (MLP):
\begin{equation}
\begin{aligned}
\hat{\mathbf{s}}[t] = \text{clamp}\Big(&
\tilde{\mathbf{y}}[t]
+ \mathbf{W}_2 \text{GELU}\big(
\mathbf{W}_1 \text{LayerNorm}(\mathbf{m}_t) + \mathbf{b}_1
\big) \\
&+ \mathbf{b}_2,\; 0,\; 1 \Big),
\end{aligned}
\label{eq:refine_mamba}
\end{equation}

where $\mathbf{W}_2, \mathbf{b}_2$ are initialized to zero. This ensures that StarBOA begins training as an exact identity mapping over the physical LIHT reconstruction, incrementally learning kinematic trajectory corrections.

\subsection{Multi-Objective Loss Function}
The network is trained end-to-end using a composite reconstruction objective:
\begin{equation}
\mathcal{L} = \|\hat{\mathbf{S}} - \mathbf{S}_{\text{GT}}\|_1 + 0.5 \|\hat{\mathbf{S}} - \mathbf{S}_{\text{GT}}\|_F^2 + 0.5 \mathcal{L}_{\text{SSIM}}(\hat{\mathbf{S}}, \mathbf{S}_{\text{GT}}),
\label{eq:loss}
\end{equation}
where $\mathcal{L}_{\text{SSIM}}$ utilizes an accelerated $7 \times 7$ pool during backpropagation, while testing evaluates exact canonical SSIM over an $11 \times 11$ Gaussian kernel ($\sigma = 1.5$).

\section{Experimental Results and Discussion}
\label{sec:results}

\begin{table}[t]
\centering
\caption{Decoupled Front-End Benchmark on Frozen LIHT Under $50\%$, $75\%$, and $90\%$ Sparsity on the DISC Dataset.}
\label{tab:frozen_benchmark}
\setlength{\tabcolsep}{2.0pt}
\begin{tabular}{@{}llccccc@{}}
\toprule
\textbf{Spar.} & \textbf{Model} & \textbf{SSIM} $\uparrow$ & $\bm{s}$ & $\bm{l}$ & $\bm{c}$ & \textbf{RMSE} $\downarrow$ \\
\midrule
\multirow{2}{*}{\textbf{50\%}} & STAR-Attn \cite{mazzieri2024attention} & 0.8595 & 0.9245 & 0.9461 & 0.9400 & 0.0849 \\
 & \textbf{StarBOA (Ours)} & \textbf{0.8822} & \textbf{0.9394} & \textbf{0.9565} & \textbf{0.9531} & \textbf{0.0746} \\
\midrule
\multirow{2}{*}{\textbf{75\%}} & STAR-Attn \cite{mazzieri2024attention} & 0.8204 & 0.8997 & 0.9289 & 0.9240 & 0.0950 \\
 & \textbf{StarBOA (Ours)} & \textbf{0.8536} & \textbf{0.9200} & \textbf{0.9427} & \textbf{0.9433} & \textbf{0.0868} \\
\midrule
\multirow{2}{*}{\textbf{90\%}} & STAR-Attn \cite{mazzieri2024attention} & 0.7331 & 0.8661 & 0.8538 & 0.8504 & 0.1340 \\
 & \textbf{StarBOA (Ours)} & \textbf{0.8047} & \textbf{0.8889} & \textbf{0.9137} & \textbf{0.9210} & \textbf{0.1107} \\
\bottomrule
\multicolumn{7}{@{}l}{\footnotesize $s$: Structure, $l$: Luminance, $c$: Contrast components of SSIM \cite{wang2004image}.}
\end{tabular}
\end{table}

\subsection{Experimental Setup}
We evaluate our framework on the public \textbf{DISC mmWave radar dataset} \cite{disc_dataset, mazzieri2024attention}, recorded using $60\text{ GHz}$ IEEE 802.11ay CIR transceivers across four human activities: \texttt{WALKING}, \texttt{RUNNING}, \texttt{SITTING}, and \texttt{HANDS}. Frames consist of $K=64$ slow-time samples per window shifted by $\delta=32$ ($T_c = 0.27\text{ ms}$, physical window duration $T_w = K T_c = 17.28\text{ ms}$, window stride $T_{\text{stride}} = \delta T_c = 8.64\text{ ms}$). Compressive sensing is tested at $50\%$, $75\%$, and $90\%$ missing chirp ratios under identical data splits, with all models trained for 20 epochs.

\subsection{Decoupled Front-End Benchmark Comparison}
To measure the exact block-level contribution of the temporal refinement mechanism, we conduct a decoupled ceteris-paribus ablation where the physical LIHT front-end (Block~A) is frozen, feeding identical coarse inputs to both refinement architectures.

As documented in Table~\ref{tab:frozen_benchmark}, fidelity is quantified via RMSE and SSIM, decomposed as $\text{SSIM}(\mathbf{x}, \mathbf{y}) = [l(\mathbf{x}, \mathbf{y})] \cdot [c(\mathbf{x}, \mathbf{y})] \cdot [s(\mathbf{x}, \mathbf{y})]$ \cite{wang2004image}, where $l(\cdot)$, $c(\cdot)$, $s(\cdot)$ denote luminance, contrast, and structure (kinematic trajectory correlation). This decomposition ensures high fidelity reflects genuine gait-curve recovery rather than trivial energy scaling. Across every sparsity step, StarBOA delivers superior performance, improving SSIM by $+2.64\%$ at $50\%$, $+4.05\%$ at $75\%$, and $+9.77\%$ at $90\%$ sparsity, while locking kinematic structure correlation above $s > 0.888$.

\begin{figure*}[t]
\centering
\includegraphics[width=0.90\textwidth]{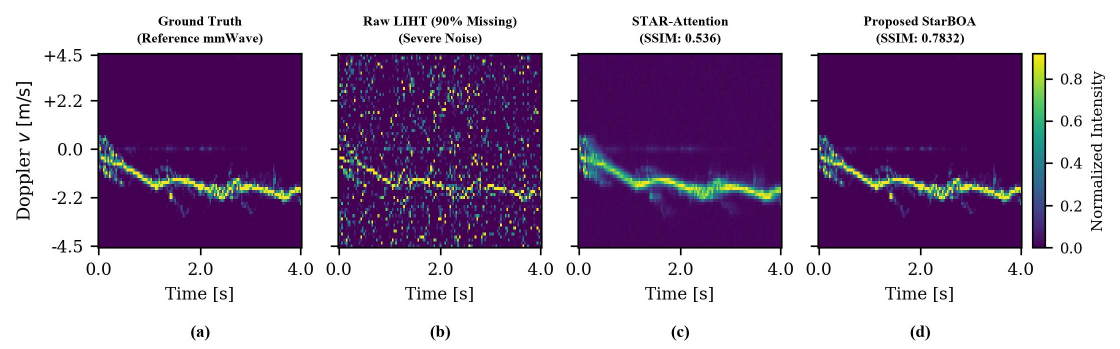}
\caption{Visual reconstruction under $90\%$ extreme sparsity on the DISC mmWave benchmark ($v \in [-4.5, 4.5]\text{ m/s}$, $t \in [0, 4]\text{ s}$). (a) Ground Truth reference mmWave spectrogram. (b) Raw LIHT input under $90\%$ pulse dropouts (severe noise and fragmented velocity profile). (c) STAR-Attention \cite{mazzieri2024attention} ($\text{SSIM} = 0.536$, diffuse moving-average smearing and blurred limb arcs). (d) Proposed StarBOA ($\text{SSIM} = \mathbf{0.7832}$, sharp torso line, continuous limb oscillation arcs, and high-contrast gait trajectory).}
\label{fig:spectrograms}
\end{figure*}

\subsection{Real-Time Streaming and Recurrent Memory Horizon}
Because StarBOA operates causally in online streaming mode, we evaluate its recovery fidelity as a function of the recurrent memory horizon $H \in \{2, \dots, 128\}$ past windows.

\begin{table}[t]
\centering
\caption{Reconstruction Fidelity vs. Recurrent Memory Horizon ($H$) at $90\%$ Chirp Sparsity.}
\label{tab:horizon}
\setlength{\tabcolsep}{3.5pt}
\begin{tabular}{@{}ccccc@{}}
\toprule
\textbf{Horizon} $\bm{H}$ & \textbf{Duration} & \textbf{SSIM} $\uparrow$ & \textbf{Struct} ($\bm{s}$) & \textbf{RMSE} $\downarrow$ \\
\midrule
$H = 2$   & $17.3\text{ ms}$   & 0.6861 & 0.8238 & 0.1457 \\
$H = 32$  & $276.5\text{ ms}$  & 0.7560 & 0.8557 & 0.1263 \\
$H = 128$ & $1105.9\text{ ms}$ & \textbf{0.7654} & \textbf{0.8601} & \textbf{0.1233} \\
\bottomrule
\end{tabular}
\end{table}

As shown in Table~\ref{tab:horizon}, recovery precision ascends monotonically as the memory horizon expands, smoothly approaching steady-state performance as the context encompasses a full human gait cycle ($H=128$, spanning $1.11\text{ s} = 128 \times 8.64\text{ ms}$). Crucially, total end-to-end inference latency per radar window is only $1.5255\text{ ms}$ ($1525.5\ \mu\text{s}$) in pure software ($730.6\ \mu\text{s}$ LIHT, $670.0\ \mu\text{s}$ Mamba step, and $124.9\ \mu\text{s}$ refinement). Operating $>11\times$ faster than the $17.28\text{ ms}$ physical window duration (and $>5\times$ faster than the $8.64\text{ ms}$ arrival stride) with strictly $0.00\text{ ms}$ lookahead, StarBOA provides ample computational headroom for real-time streaming on edge ISAC processors.

\subsection{Root-Cause Analysis: Attention Entropy Collapse vs. Recurrent State Memory}
To uncover the structural mechanism governing performance under extreme subsampling, Table~\ref{tab:full_pipeline}(b) analyzes the historical context distribution under $90\%$ chirp missingness.

In STAR, candidate $\tilde{\mathbf{y}}[t]$ is matched against $N_p=6$ past windows via dot products. Under $90\%$ missingness, severe noise flattens attention weights across the 6 historical windows to near-identical values ($\alpha \approx 0.172\text{--}0.175$). Empirical evaluation reveals that attention entropy reaches $H = 2.5840\text{ bits}$, matching the theoretical maximum $H_{\text{uniform}} = \log_2(6) = 2.5850\text{ bits}$. This entropy collapse compounds STAR's short $52\text{ ms}$ buffer limitation, explaining why attention degrades to moving-average blur while recurrent state-space models remain robust across all sparsity tiers (Table~\ref{tab:full_pipeline}(a)).

\textbf{Why Recurrent State Spaces Prevent Collapse:} Unlike attention—which re-evaluates a noisy instantaneous query $\tilde{\mathbf{y}}[t]$ against an external buffer of corrupted past estimates—StarBOA maintains an internal recurrent hidden state $\mathbf{H}_t \in \R^{D \times N}$. Because state transitions are governed by continuous recurrence and selective input projections ($\Delta_t, \mathbf{B}_t$), the state does not depend on static window averaging. Instead, instantaneous noise bursts are naturally suppressed by the selective state transition, while coherent temporal trajectories are integrated across hundreds of slow-time windows ($H \ge 128$). Consequently, StarBOA remains robust by construction under severe pulse missingness.
\subsection{Visual Spectrogram Inspection}
Fig.~\ref{fig:spectrograms} visually corroborates Table~\ref{tab:frozen_benchmark}: StarBOA reconstructs a continuous torso trajectory and limb-oscillation arcs where STAR-Attention exhibits moving-average smearing and fragmented footfalls (per-panel SSIM in caption).

\begin{table}[t]
\centering
\caption{Full End-to-End Pipeline Comparison and STAR Attention Distribution Under $90\%$ Sparsity.}
\label{tab:full_pipeline}
\setlength{\tabcolsep}{2.5pt}
\begin{tabular}{@{}llccc@{}}
\toprule
\multicolumn{5}{@{}c@{}}{\textbf{(a) End-to-End Pipeline Reconstruction Across Sparsity Tiers}} \\
\midrule
\textbf{Sparsity} & \textbf{Architecture} & \textbf{RMSE} $\downarrow$ & \textbf{SSIM} $\uparrow$ & $\bm{\Delta}$\textbf{SSIM} \\
\midrule
\multirow{2}{*}{\textbf{50\%}} & STAR \cite{mazzieri2024attention} (reported) & 0.0545 & 0.8840 & Reference \\
 & \textbf{StarBOA (Ours)} & \textbf{0.0506} & \textbf{0.9219} & $\mathbf{+0.0379}$ \\
\midrule
\multirow{2}{*}{\textbf{75\%}} & STAR \cite{mazzieri2024attention} (reported) & 0.0779 & 0.7450 & Reference \\
 & \textbf{StarBOA (Ours)} & \textbf{0.0772} & \textbf{0.8679} & $\mathbf{+0.1229}$ \\
\midrule
\multirow{2}{*}{\textbf{90\%}} & STAR \cite{mazzieri2024attention} (reported) & 0.1213 & 0.5360 & Reference \\
 & \textbf{StarBOA (Ours)} & \textbf{0.1211} & \textbf{0.7832} & $\mathbf{+0.2472}$ \\
\midrule
\multicolumn{5}{@{}c@{}}{\textbf{(b) STAR Attention Weights ($90\%$ Sparsity, $N_p=6$)}} \\
\midrule
\multicolumn{5}{@{}l}{\textbf{Weights:} $\alpha_{t-1}=0.174, \; \alpha_{t-2}=0.173, \; \alpha_{t-3}=0.175,$} \\
\multicolumn{5}{@{}l}{\phantom{\textbf{Weights:}} $\alpha_{t-4}=0.174, \; \alpha_{t-5}=0.172, \; \alpha_{t-6}=0.174$} \\
\multicolumn{5}{@{}l}{Uniform: $1/6 \approx 0.167$ \quad $\vert$ \quad Entropy: $H = 2.5840\text{ bits} \approx \log_2(6)$} \\
\bottomrule
\end{tabular}
\end{table}

\subsection{Full End-to-End Pipeline Evaluation}
Table~\ref{tab:full_pipeline}(a) compares our full end-to-end StarBOA against STAR's officially reported numbers \cite{mazzieri2024attention}: the SSIM gain widens from $+0.0379$ at $50\%$ to $+0.1229$ at $75\%$ and $+0.2472$ at $90\%$ sparsity, consistent with the compounding buffer-span and entropy-collapse mechanisms above (the controlled, same-environment ablation isolating this effect is given in Table~\ref{tab:frozen_benchmark}). This benefits ISAC design directly: replacing fragile attention with state-space unrolling lets base stations cut radar transmission to a $10\%$ duty cycle without compromising HAR-critical sensing fidelity.

\section{Conclusion}
\label{sec:conclusion}
In this letter, we showed that non-parametric attention in unrolled radar recovery suffers from a short $52\text{ ms}$ buffer and collapses to maximum uniform entropy ($H{=}2.584\text{ bits}{\approx}\log_2 6$) under severe missingness. We proposed StarBOA, replacing attention with a causal continuous-time Mamba state space that captures context across a full $1.1\text{ s}$ gait cycle with $O(1)$ memory. On the DISC mmWave benchmark, StarBOA consistently outperforms STAR across all sparsity tiers, with SSIM gains widening to $+0.2472$ at $90\%$ missing chirps while executing in $1.53\text{ ms}$ with zero lookahead. This allows base stations to allocate $90\%$ of resources to communications while sustaining high-fidelity micro-Doppler sensing.

\bibliographystyle{IEEEtran}

\end{document}